\documentclass{article} 
\usepackage{colm2024_conference}
\usepackage[misc]{ifsym}
\usepackage{microtype}
\usepackage{hyperref}
\usepackage{url}
\usepackage{booktabs}
\definecolor{darkblue}{rgb}{0, 0, 0.5}
\hypersetup{colorlinks=true, citecolor=darkblue, linkcolor=darkblue, urlcolor=darkblue}
\usepackage{multirow}
\usepackage{booktabs}
\usepackage{multicol}
\usepackage{appendix}
\usepackage{graphicx}
\usepackage{amsmath}
\usepackage{amssymb}
\usepackage{amsfonts}
\usepackage{amsthm}
\definecolor{authorcolor}{RGB}{18, 101, 251}
\usepackage{cleveref}
\usepackage{booktabs}
\usepackage[table]{xcolor}
\usepackage[svgnames,dvipsnames]{xcolor}  
\usepackage{graphicx}   
\usepackage{subcaption}  
\usepackage{enumitem}
\usepackage{algorithm}
\usepackage{algorithmicx}  
\usepackage{algpseudocode}  
\title{UNSEEN: Enhancing Dataset Pruning from a Generalization Perspective}

\author{
    \small
  {\bf
    Furui Xu $^*$$^{\spadesuit}$
    Shaobo Wang $^*$$^{\spadesuit,\clubsuit}$
    Jiajun Zhang $^{\spadesuit}$
    Chenghao Sun $^{\spadesuit}$
    Haixiang Tang $^{\spadesuit}$  
    Linfeng Zhang $^{\text{\Letter}\spadesuit}$
    \vspace{3pt}
  } \\
    \small
    {
    $^{\spadesuit}$ EPIC Lab, SJTU $\quad$
    $^{\clubsuit}$ Alibaba Group $\quad$
  $^*$ Equal contribution $\quad$ 
    $^{\text{\Letter}}$ Corresponding author
    }
}

\colmfinalcopy 
\begin{document}

\maketitle

\begin{abstract}
The growing scale of datasets in deep learning has introduced significant computational challenges. Dataset pruning addresses this challenge by constructing a compact but informative coreset from the full dataset with comparable performance. Previous approaches typically establish scoring metrics based on specific criteria to identify representative samples. However, these methods predominantly rely on sample scores obtained from the model's performance during the training (i.e., fitting) phase. As scoring models achieve near-optimal performance on training data, such fitting-centric approaches induce a dense distribution of sample scores within a narrow numerical range. This concentration reduces the distinction between samples and hinders effective selection. To address this challenge, we conduct dataset pruning from the perspective of generalization, i.e., scoring samples based on models not exposed to them during training. We propose a plug-and-play framework, UNSEEN, which can be integrated into existing dataset pruning methods. Additionally, conventional score-based methods are single-step and rely on models trained solely on the complete dataset, providing limited perspective on the importance of samples. To address this limitation, we scale UNSEEN to multi-step scenarios and propose an incremental selection technique through scoring models trained on varying coresets, and optimize the quality of the coreset dynamically. Extensive experiments demonstrate that our method significantly outperforms existing state-of-the-art (SOTA) methods on CIFAR-10, CIFAR-100, and ImageNet-1K. Notably, on ImageNet-1K,  UNSEEN achieves lossless performance while reducing training data by 30\%.     
\end{abstract}

\section{Introduction}

\begin{figure}[tb!]
  \centering
  \begin{subfigure}{.8\linewidth} 
    \includegraphics[width=.8\linewidth]{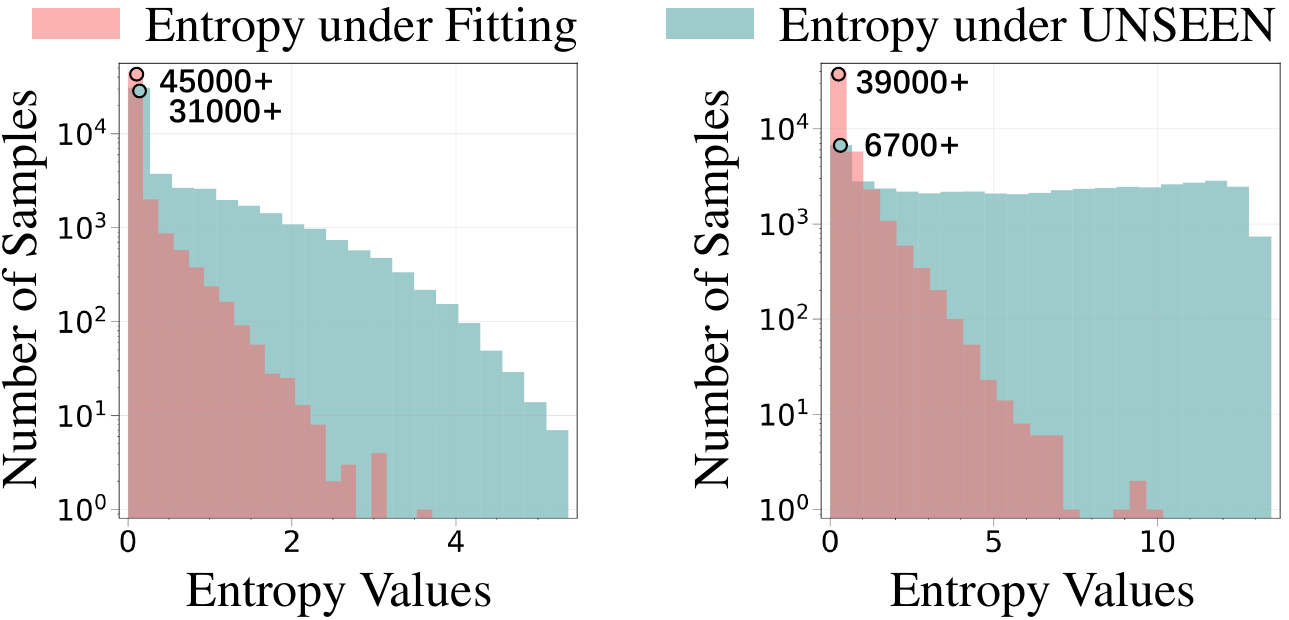} 
    \caption{Entropy distribution under the Fitting and UNSEEN frameworks on CIFAR-10 (left) and CIFAR-100 (right) datasets.}
    \label{Entropy Distribution}
  \end{subfigure}
  \centering
  \begin{subfigure}{.8\linewidth} 
    \includegraphics[width=.8\linewidth]{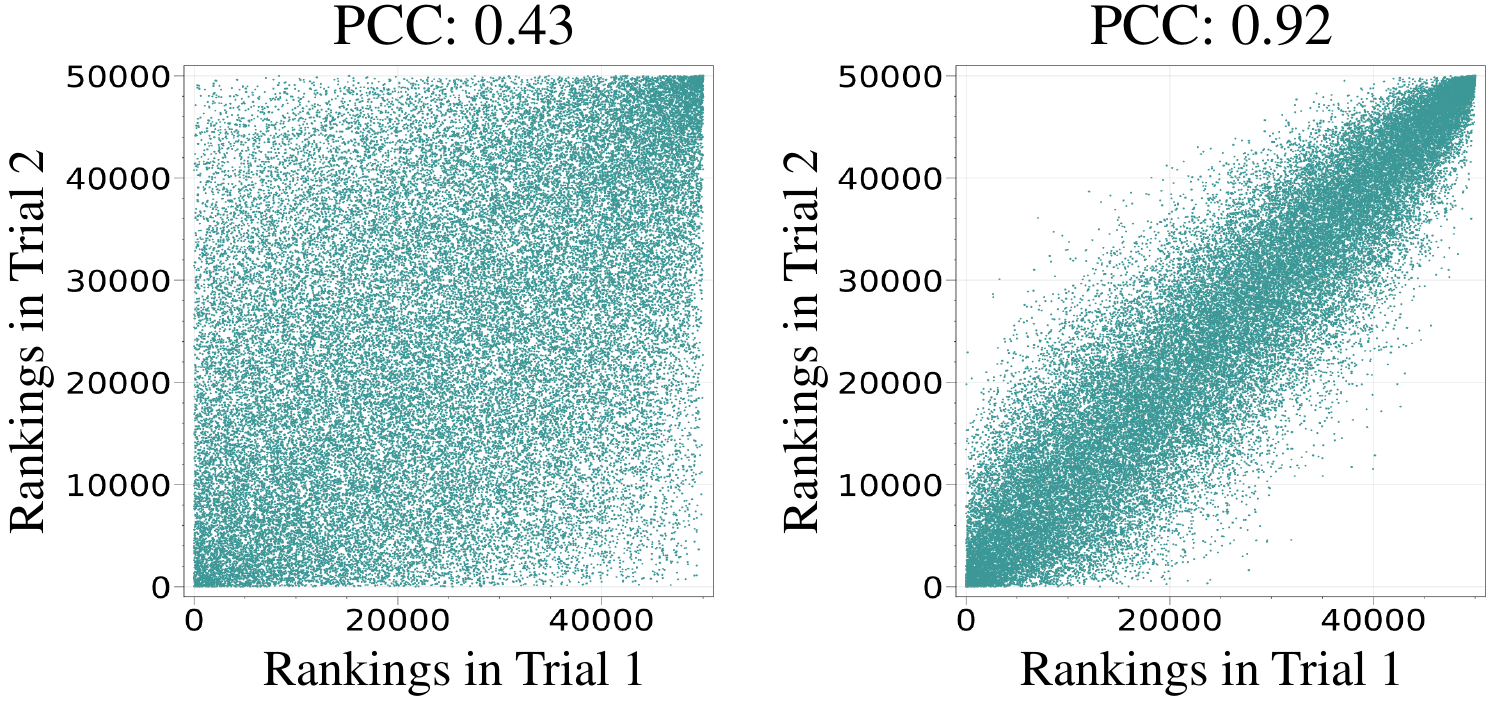}
    \caption{Rank distribution of two identical trials with different random seeds under the Fitting (Top) and UNSEEN (Bottom) frameworks.}
    \label{rank comparison}
  \end{subfigure}
  
  \caption{(a) Distribution of Entropy score on CIFAR-10 and CIFAR-100 under the fitting and UNSEEN frameworks. Under the fitting framework, Entropy scores exhibit dense clustering. Conversely, UNSEEN achieves uniform score dispersion, substantially improving discriminative separability. (b) Distribution of the rank assigned to each sample in the overall score ranking in two identical CIFAR-100 trials with different random seeds. Sample ranks fluctuate significantly under fitting but remain stable under UNSEEN. The Pearson correlation coefficient (PCC) between trials is 0.92 for UNSEEN, much higher than 0.43 under fitting.}
  
  \label{Intro_fig}
\end{figure}

\begin{figure*}[tb!]
  \centering
  \includegraphics[width=0.9\linewidth]{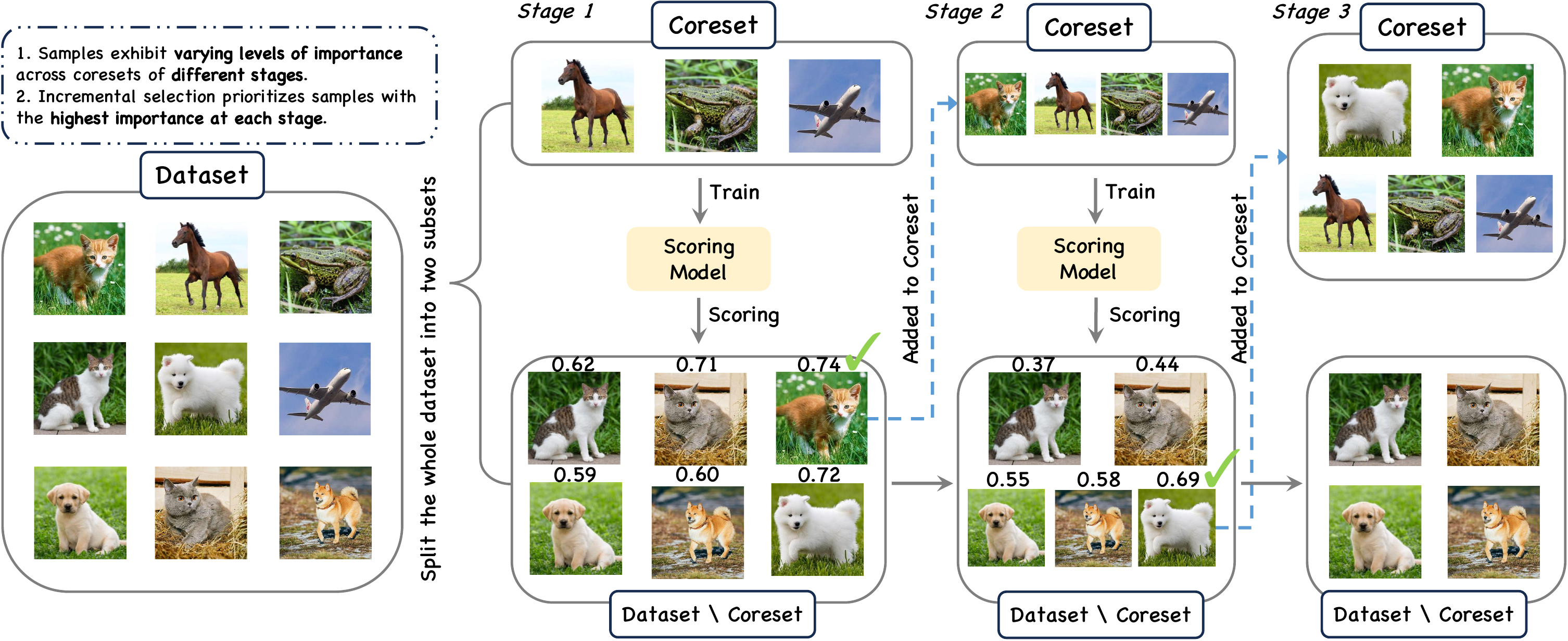}

  \caption{Samples exhibit varying levels of importance across coresets of different stages. Incremental selection prioritizes samples with the highest importance at each stage,  offering a more principled and adaptive approach to coreset construction.}
  \label{incremental}
\end{figure*}

In recent decades, the rapid development of deep learning has been driven by large-scale datasets~\citep{deng2009imagenet,kuznetsova2020open,zhou2017places}, producing many amazing achievements ~\citep{floridi2020gpt,chowdhery2023palm,radford2021learning}. However, this approach is inherently costly and requires substantial computational resources and considerable time~\citep{cazenavette2022dataset,yu2023dataset,zhao2020dataset}. Moreover, a significant portion of the data is redundant or erroneous~\citep{zhang2024spanning,zheng2022coverage,xia2022moderate,paul2021deep}, contributing minimally to the improvement of model performance. Dataset pruning~\citep{guo2022deepcore}, also known as coreset selection, mitigates data redundancy by constructing a compact subset that enables the model to achieve performance comparable to that obtained with the full dataset. This strategy emphasizes identifying and selecting the most informative and representative samples.

Dataset pruning methods commonly identify important samples through various strategies, including geometry features~\citep{xia2022moderate, zheng2022coverage,  wan2024contributing}, uncertainty~\citep{coleman2019selection, pleiss2020identifying, he2024large, zhang2024spanning}, error~\citep{paul2021deep,toneva2018an}, and decision boundary~\citep{margatina2021active,yang2024mind}. 
These methods are based on their scores derived from the model's performance during the training (\emph{i.e.,} fitting) phase. 
However, as models exhibit a strong capacity for sample fitting, the score distribution becomes densely concentrated. As demonstrated in ~\Cref{Entropy Distribution}, 
Under the conventional fitting framework, more than 90\% of CIFAR-10 samples and 78\% of CIFAR-100 samples exhibit Entropy values close to zero. Due to the overly clustered score distribution of samples, the model frequently fails to accurately identify challenging samples, and the selection results exhibit severe instability. As shown in \Cref{rank comparison}, the rank assigned to each sample in the overall score ranking varies significantly between two otherwise identical experiments initialized with different random seeds. This observation indicates that conventional methods under the fitting framework exhibit limitations in effectively and robustly differentiating samples of varying difficulty.

To overcome this challenge, we propose a novel framework called \textbf{UNSEEN}, designed to assess the importance of data samples from a generalization perspective. Specifically, we employ cross-validation to facilitate mutual scoring across multiple models, where each sample's importance is determined exclusively by models that never encountered it during training. 
The implementation involves randomly partitioning the full dataset into mutually exclusive folds of equal size, training scoring models on each fold, and subsequently utilizing these models to score the samples excluded from their respective folds. As shown in \Cref{Entropy Distribution}, our generalization-based framework UNSEEN produces Entropy scores with more uniform dispersion across a broader scoring range compared to the conventional fitting-based approach. \Cref{rank comparison} demonstrates that our method significantly enhances robust and discriminative selection.

Additionally, existing score-based methods are single-step \emph{i.e.,} score samples once with the scoring model trained solely on the full dataset. As demonstrated in \Cref{incremental}, samples exhibit varying levels of importance when scored by models trained on different coresets. We frame coreset construction as an incremental process, considering the difficulty of samples for models trained on coresets of varying sizes. To comprehensively assess the importance of samples, we scale UNSEEN to multi-step scenarios and propose an evaluate-and-refill paradigm,   \textbf{incremental selection}. It initiates by constructing an initial coreset through a selection criterion under the fitting or UNSEEN framework. A scoring model is subsequently trained on the coreset and is then employed to score pruned samples. Those that obtain the highest scores are incorporated into the coreset. This procedure persists until the final coreset attains the target cardinality. 

Our contributions in this paper are as follows:

\begin{itemize}[leftmargin=10pt, topsep=0pt, itemsep=1pt, partopsep=1pt, parsep=1pt] 
    \item We expose the limitations of previous fitting-based methods and introduce UNSEEN, a plug-and-play framework designed from a generalization perspective.
    \item We enhance conventional single-step pruning methods by scaling UNSEEN to a multi-step selection process, and propose incremental selection (IS), which provides a more comprehensive assessment of sample importance.
    \item Our method outperforms existing SOTA methods on CIFAR-10, CIFAR-100, and ImageNet-1K, and achieves 30\% lossless pruning on ImageNet-1K.
    \item We extend the notion of difficulty from individual samples to the class level, prioritizing the minimization of inter-class disparity over uniform treatment across categories.

\end{itemize}

\begin{figure*}[tb!]
    \centering
    \includegraphics[width=0.9\linewidth]{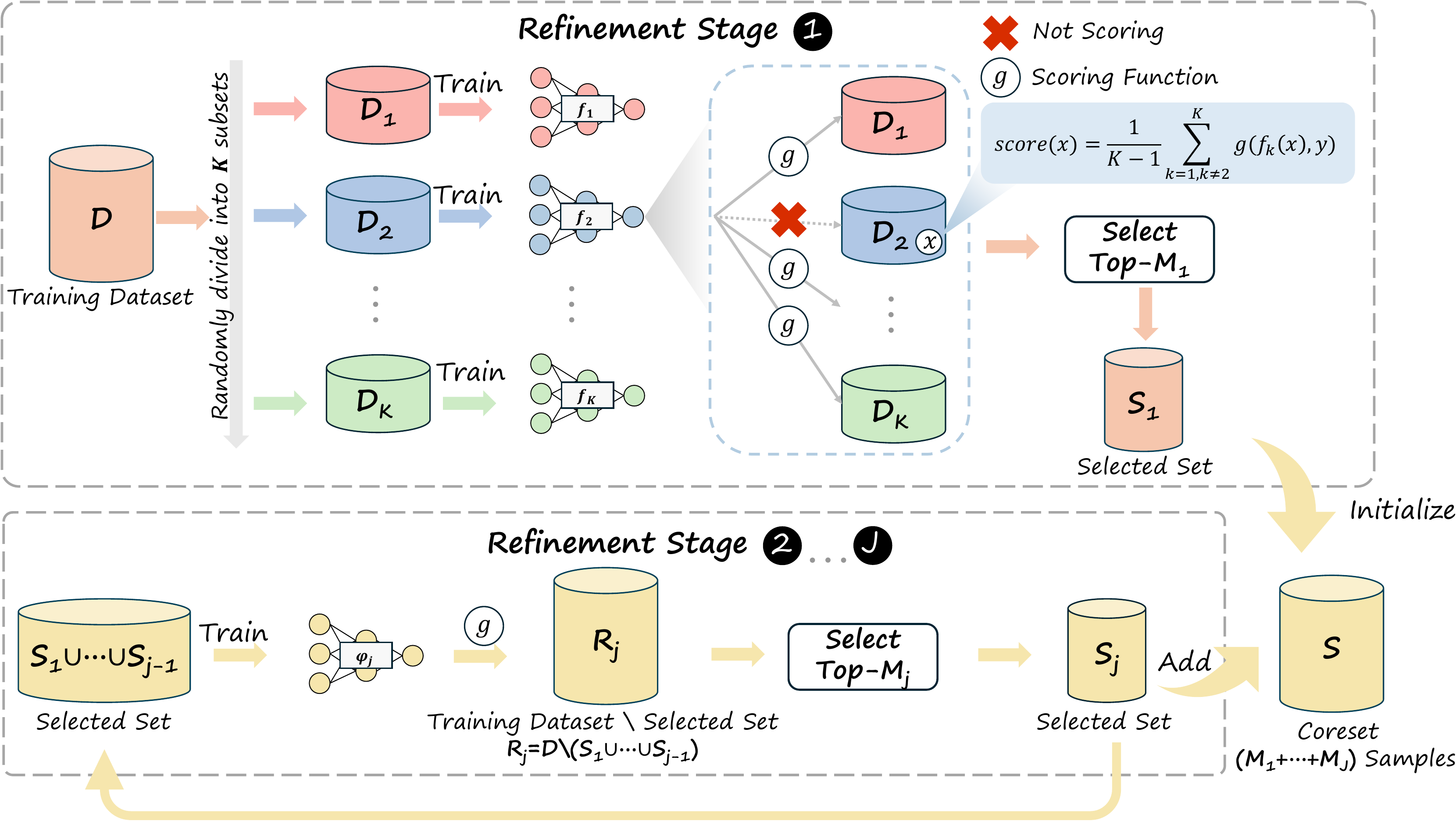}

    \caption{The pipeline of UNSEEN. First, the training dataset is randomly partitioned into $K$ equal-sized subsets. Then, for each subset, a scoring model is trained and used to assign scores to the samples in the complementary subsets. The top $M_1$ samples with the highest scores are selected to form the initial coreset $S_1$. Next, a scoring model is trained on the selected samples and used to score the remaining unselected samples. Samples with the highest scores are incrementally added to the coreset. This procedure is repeated until the desired number of samples has been selected.}
    \label{fig:pipeline}
\end{figure*}

\section{Related work}
\label{sec: formatting}

\noindent \textbf{Dataset Pruning.} Current approaches to dataset pruning mainly include score-based and optimization-based methods. Emerging as the predominant paradigm, score-based methods select the representative subset by scoring samples based on specific metrics.    Various criteria have been developed, such as geometry-based~\citep{welling2009herding, xia2022moderate, zheng2022coverage}, uncertainty-based~\citep{coleman2019selection, pleiss2020identifying, he2024large,zhang2024spanning}, error-based~\citep{paul2021deep, toneva2018an}, decision-boundary-based~\citep{margatina2021active,yang2024mind}.  (I) \textit{Geometry}-based methods leverage spatial distribution characteristics to identify representative samples. Moderate~\citep{xia2022moderate} employs median distance criteria, retaining samples near the distributional median. CCS~\citep{zheng2022coverage} selects samples that ensure data coverage. (II) \textit{Uncertainty}-based methods focus on identifying samples with low confidence or high uncertainty of prediction. Entropy~\citep{coleman2019selection} selects samples with elevated cross-entropy. DynUnc~\citep{he2024large} prioritizes samples with high uncertain prediction. TDDS~\citep{zhang2024spanning}
targets samples with larger projected gradient variances. (III) \textit{Error}-based methods identify and remove data points that demonstrate minimal contributions to model performance. Forgetting~\citep{toneva2018an} retains frequently misclassified samples that exhibit persistent training errors throughout the learning process. Optimization-based methods treat dataset pruning as an optimization problem. Glister~\citep{killamsetty2021glister} introduces validation data on the outer optimization and the log-likelihood in the bilevel optimization. While theoretically promising, these methods face practical implementation barriers due to intricate bilevel optimization~\citep{yang2024mind,wang2025winning}. Additionally, proxy-based methods~\citep{coleman2019selection,sachdeva2021svp,wang2025datawhisperer} use lightweight or shallow models to fit the training dataset, thus reducing computational cost.

\noindent \textbf{Dataset Synthesis and Distillation.} A parallel and increasingly prominent line of research moves beyond merely filtering existing datasets and instead focuses on actively transforming or generating new, higher-quality data. This paradigm, often termed \textit{dataset synthesis} or \textit{distillation}, aims to engineer a more informative and robust training signal than what is available in the raw data distribution~\citep{wang2025socratic,wang2024samples,wang2025drupi,liu2025shifting,min2025imagebinddc,huang2025r}.

\section{Methodology}
\subsection{Preliminaries}
Consider a classification task with training dataset $\mathcal{D} = \{(x_i, y_i)\}_{i=1}^N$ where $x_i \in \mathbb{R}^d$ denotes input features and $y_i \in \{1,\ldots,c\}$ represents class labels for a $c$-category problem. The data follows an unknown distribution $\mathcal{P}$, and we aim to train a neural network $f_\theta: \mathbb{R}^d \rightarrow \mathbb{R}^c$ parameterized by $\theta \in \mathbb{R}^m$. The model minimizes the empirical risk $\mathcal{L}(\mathcal{D}; \theta) = \frac{1}{N}\sum_{i=1}^N \ell(f_\theta(x_i), y_i)$, where $\ell: \mathbb{R}^c \times \{1,\ldots,c\} \rightarrow \mathbb{R}^+$ is a loss function such as cross-entropy. The dataset pruning objective seeks a coreset $\mathcal{S} \subset \mathcal{D}$ with cardinality $|\mathcal{S}| = M < N$ that preserves model generalization performance. Formally, we require:
\begin{equation}
    \underset{\substack{(x,y)\sim\mathcal{P}\\ \theta_0\sim\mathcal{P}_{\theta_0}}}{\mathbb{E}} \left[\ell(f_{\theta_{\mathcal{D}}}(x), y)\right] \approx \underset{\substack{(x,y)\sim\mathcal{P}\\ \theta_0\sim\mathcal{P}_{\theta_0}}}{\mathbb{E}} \left[\ell(f_{\theta_{\mathcal{S}}}(x), y)\right],
\end{equation}
where $\theta_{\mathcal{D}}$ and $\theta_{\mathcal{S}}$ denote parameters trained on $\mathcal{D}$ and $\mathcal{S}$ respectively, with initialization $\theta_0 \sim \mathcal{P}_{\theta_0}$.

\begin{algorithm}[tb!]
\caption{UNSEEN-Incremental Selection}
\label{algorithm1}
\begin{algorithmic}[1]
\Require Training dataset \(\mathcal{D}\) with \(N\) samples, target pruning rate \(p \in (0,1)\), number of partitions \(K\), number of refinement stages \(J\)
\State Compute target coreset size: \( M \gets \lfloor N(1-p) \rfloor \)
\State Define per-stage selection sizes \( \{M_j\}_{j=1}^{J} \) such that \( \sum_{j=1}^{J} M_j = M \)
\Statex \Comment{\textbf{Cross-Validated UNSEEN Scoring}}
\State Partition \(\mathcal{D}\) into \(K\) mutually exclusive subsets: \( \{\mathcal{D}_k\}_{k=1}^K \), ensuring \( \mathcal{D}_i \cap \mathcal{D}_j = \emptyset \) and \( \bigcup_{k=1}^{K} \mathcal{D}_k = \mathcal{D} \)
\For{\( k = 1 \) to \( K \)}
    \State Train model \(f_{\theta_k}\) on \(\mathcal{D}_k\)
    \For{each sample \( (x_i, y_i) \in \mathcal{D}_k^c = \mathcal{D} \setminus \mathcal{D}_k \)}
        \State   
        \(
        s(x_i) = \ell(f_{\theta_k}(x_i), y_i)
        \)\Comment{Compute score}
    \EndFor
\EndFor
\For{each sample \( x_i \in \mathcal{D} \)}
    \State   
    \(
    \tilde{s}(x_i) = \frac{s(x_i)}{\sum_{j=1}^{N} s(x_j)}
    \)\Comment{Normalize scores}
\EndFor
\State Initialize coreset:  
\(
\mathcal{S}_1 \gets \textbf{Select}(\{\tilde{s}(x_i)\}_{x_i \in \mathcal{D}}, M_1)
\)
\State Define pruned dataset: \( \mathcal{R}_1 \gets \mathcal{D} \setminus \mathcal{S}_1 \)

\Statex \Comment{\textbf{Incremental Selection}}
\For{\( j = 2 \) to \( J \)}
    \State Train model \(f_{\theta_j}\) on \(\mathcal{S}_{j-1}\)
    \For{each sample \( (x_i, y_i) \in \mathcal{R}_{j-1} \)}
        \State   
        \(
        s_j(x_i) = \ell(f_{\theta_j}(x_i), y_i)
        \)
    \EndFor
    \State   
    \(
    \Delta_j \gets \textbf{Select}(\{s_j(x_i)\}_{(x_i,y_i) \in \mathcal{R}_{j-1}}, M_j)
    \)\Comment{Select top \( M_j \) samples}
    \State Update coreset:  
    \(
    \mathcal{S}_j \gets \mathcal{S}_{j-1} \cup \Delta_j
    \)
    \State Update pruned dataset:  
    \(
    \mathcal{R}_j \gets \mathcal{R}_{j-1} \setminus \Delta_j
    \)
\EndFor
\State \Return Final coreset \( \mathcal{S}_J \) and final model \( f_{\theta_J} \)
\end{algorithmic}
\end{algorithm}

\subsection{UNSEEN}

We implement UNSEEN by cross-validated sample scoring and scale UNSEEN to incremental selection. The detailed pseudo-code of our approach is presented in \Cref{algorithm1}.

\noindent \textbf{Cross-validated UNSEEN Scoring}. The process initiates by partitioning the original dataset $\mathcal{D}$ into $K$ mutually exclusive subsets $\{\mathcal{D}_k\}_{k=1}^K$ and corresponding complements $\mathcal{D}_k^c =\mathcal{D} \setminus \mathcal{D}_k$ through uniform random sampling, ensuring $\mathcal{D}_i \cap \mathcal{D}_j = \emptyset$ for all $i \neq j$ while maintaining $\bigcup_{k=1}^K \mathcal{D}_k = \mathcal{D}$. For each partition $\mathcal{D}_k$, a neural network $f_{\theta_k}$ is trained by optimizing $\theta_k = \arg\min_\theta \mathcal{L}(\mathcal{D}_k; \theta)$. This network subsequently generates scores on the corresponding complement $s(x_i) = \ell(f_{\theta_k}(x_i), y_i)$, $(x_i, y_i) \in \mathcal{D}_k^c$, establishing a cross-validated assessment where models exclusively evaluate samples excluded from their training scenario. We adopt the basic Entropy score as the scoring function.

\noindent \textbf{Incremental Selection (IS)}. Given a target pruning rate $p \in (0,1)$, the algorithm selects $M = \lfloor N(1-p) \rfloor$ samples through an incremental refinement process over $J$ stages. It begins with score normalization $\tilde{s}(x_i) = s(x_i)/\sum_{j=1}^N s(x_j)$ and sets per-stage selection sizes $M_j$ such that $\sum_{j=1}^J M_j = M$. At stage $j$, a scoring network is trained on the current coreset $\mathcal{S}{j-1}$ and assigns loss scores $s_j(x_i) = \ell(f{\theta_j}(x_i), y_i)$ to samples in the remaining set $\mathcal{R}{j-1} = \mathcal{D} \setminus \mathcal{S}{j-1}$. The top-$M_j$ samples $\Delta_j$ with highest scores are added to the coreset, i.e., $\mathcal{S}j = \mathcal{S}{j-1} \cup \Delta_j$, while the residual set is updated accordingly. This process continues until the final coreset $\mathcal{S}_J$ reaches size $M$.

\begin{table*}[tb!]
\caption{Comprehensive comparison on CIFAR-10 and CIFAR-100 datasets with ResNet-18. The accuracy on the full dataset of CIFAR-10 and CIFAR-100 is 95.50\% and 79.24\%. UNSEEN outperforms full-data training when pruning 30\% of CIFAR-10 samples. For CIFAR-100, it prunes 30\% of training samples with only a 0.63\% accuracy drop.}
\centering
\resizebox{0.99\textwidth}{!}{
\begin{tabular}{@{}c| *{5}{c} *{5}{c} @{}}
\toprule
\multicolumn{1}{c|}{Dataset} & 
\multicolumn{5}{c}{\textbf{CIFAR-10}} & \multicolumn{5}{c}{\textbf{CIFAR-100}} \\ 
\cmidrule(r){0-0}\cmidrule(lr){2-6} \cmidrule(lr){7-11}
Prune Rate & 30\% & 40\% & 50\% & 60\% & 70\% & 30\% & 40\% & 50\% & 60\% & 70\% \\ 
\cmidrule(r){0-0}\cmidrule(lr){2-6} \cmidrule(lr){7-11}

Random & 94.67 & 94.15 & 93.27 & 92.49 & 91.04 & 76.00 & 74.32 & 72.37 & 69.87 & 66.26  \\

Entropy & 94.77\textsubscript{\textcolor{PineGreen}{$\uparrow$0.10}} & 94.37\textsubscript{\textcolor{PineGreen}{$\uparrow$0.22}} & 93.87\textsubscript{\textcolor{PineGreen}{$\uparrow$0.60}} & 92.76\textsubscript{\textcolor{PineGreen}{$\uparrow$0.27}} & 90.83\textsubscript{\textcolor{OrangeRed}{$\downarrow$0.21}} & 76.73\textsubscript{\textcolor{PineGreen}{$\uparrow$0.73}} & 74.94\textsubscript{\textcolor{PineGreen}{$\uparrow$0.62}} & 72.21\textsubscript{\textcolor{OrangeRed}{$\downarrow$0.16}} & 68.41\textsubscript{\textcolor{OrangeRed}{$\downarrow$1.46}} & 62.47\textsubscript{\textcolor{OrangeRed}{$\downarrow$3.79}} \\

Margin & 94.83\textsubscript{\textcolor{PineGreen}{$\uparrow$0.16}} & 94.45\textsubscript{\textcolor{PineGreen}{$\uparrow$0.30}} & 93.75\textsubscript{\textcolor{PineGreen}{$\uparrow$0.48}} & 92.76\textsubscript{\textcolor{PineGreen}{$\uparrow$0.27}} & 90.35\textsubscript{\textcolor{OrangeRed}{$\downarrow$0.69}} & 76.64\textsubscript{\textcolor{PineGreen}{$\uparrow$0.64}} & 74.71\textsubscript{\textcolor{PineGreen}{$\uparrow$0.39}} & 71.95\textsubscript{\textcolor{OrangeRed}{$\downarrow$0.42}} & 67.96\textsubscript{\textcolor{OrangeRed}{$\downarrow$1.91}} & 62.15\textsubscript{\textcolor{OrangeRed}{$\downarrow$4.11}} \\

Least Confidence & 94.11\textsubscript{\textcolor{OrangeRed}{$\downarrow$0.56}} & 93.51\textsubscript{\textcolor{OrangeRed}{$\downarrow$0.64}} & 93.08\textsubscript{\textcolor{OrangeRed}{$\downarrow$0.19}} & 91.63\textsubscript{\textcolor{OrangeRed}{$\downarrow$0.86}} & 90.09\textsubscript{\textcolor{OrangeRed}{$\downarrow$0.95}} & 76.54\textsubscript{\textcolor{PineGreen}{$\uparrow$0.54}} & 74.79\textsubscript{\textcolor{PineGreen}{$\uparrow$0.47}} & 72.66\textsubscript{\textcolor{PineGreen}{$\uparrow$0.29}} & 69.00\textsubscript{\textcolor{OrangeRed}{$\downarrow$0.87}} & 63.92\textsubscript{\textcolor{OrangeRed}{$\downarrow$2.34}} \\

AUM & 95.49\textsubscript{\textcolor{PineGreen}{$\uparrow$0.82}} & 95.46\textsubscript{\textcolor{PineGreen}{$\uparrow$1.31}} & 95.22\textsubscript{\textcolor{PineGreen}{$\uparrow$1.95}} & 94.90\textsubscript{\textcolor{PineGreen}{$\uparrow$2.41}} & 92.44\textsubscript{\textcolor{PineGreen}{$\uparrow$1.40}} & 77.98\textsubscript{\textcolor{PineGreen}{$\uparrow$1.98}} & 75.41\textsubscript{\textcolor{PineGreen}{$\uparrow$1.09}} & 68.86\textsubscript{\textcolor{OrangeRed}{$\downarrow$3.51}} & 56.42\textsubscript{\textcolor{OrangeRed}{$\downarrow$13.45}} & 38.54\textsubscript{\textcolor{OrangeRed}{$\downarrow$27.72}} \\

EL2N & 95.36\textsubscript{\textcolor{PineGreen}{$\uparrow$0.69}} & 95.27\textsubscript{\textcolor{PineGreen}{$\uparrow$1.12}} & 95.10\textsubscript{\textcolor{PineGreen}{$\uparrow$1.83}} & 94.58\textsubscript{\textcolor{PineGreen}{$\uparrow$2.09}} & 91.10\textsubscript{\textcolor{PineGreen}{$\uparrow$0.06}} & 77.44\textsubscript{\textcolor{PineGreen}{$\uparrow$1.74}} & 74.47\textsubscript{\textcolor{PineGreen}{$\uparrow$0.15}} & 66.75\textsubscript{\textcolor{OrangeRed}{$\downarrow$5.62}} & 52.81\textsubscript{\textcolor{OrangeRed}{$\downarrow$17.06}} & 35.03\textsubscript{\textcolor{OrangeRed}{$\downarrow$31.23}} \\

Forgetting & 95.48\textsubscript{\textcolor{PineGreen}{$\uparrow$0.81}} & 95.47\textsubscript{\textcolor{PineGreen}{$\uparrow$1.32}} & 95.34\textsubscript{\textcolor{PineGreen}{$\uparrow$2.07}} & 95.03\textsubscript{\textcolor{PineGreen}{$\uparrow$2.54}} & 93.22\textsubscript{\textcolor{PineGreen}{$\uparrow$2.18}} & 78.16\textsubscript{\textcolor{PineGreen}{$\uparrow$2.16}} & 76.01\textsubscript{\textcolor{PineGreen}{$\uparrow$1.69}} & 71.74\textsubscript{\textcolor{OrangeRed}{$\downarrow$0.63}} & 62.42\textsubscript{\textcolor{OrangeRed}{$\downarrow$7.45}} & 48.48\textsubscript{\textcolor{OrangeRed}{$\downarrow$17.78}} \\

CCS & 95.45\textsubscript{\textcolor{PineGreen}{$\uparrow$0.78}} & 95.38\textsubscript{\textcolor{PineGreen}{$\uparrow$1.23}} & 95.14\textsubscript{\textcolor{PineGreen}{$\uparrow$1.87}} & 94.54\textsubscript{\textcolor{PineGreen}{$\uparrow$2.05}} & 91.77\textsubscript{\textcolor{PineGreen}{$\uparrow$0.73}} & 75.66\textsubscript{\textcolor{OrangeRed}{$\downarrow$0.34}} & 73.83\textsubscript{\textcolor{OrangeRed}{$\downarrow$0.49}} & 70.49\textsubscript{\textcolor{OrangeRed}{$\downarrow$1.88}} & 66.74\textsubscript{\textcolor{OrangeRed}{$\downarrow$3.13}} & 60.81\textsubscript{\textcolor{OrangeRed}{$\downarrow$5.45}} \\

DynUnc & 95.49\textsubscript{\textcolor{PineGreen}{$\uparrow$0.82}} & 95.45\textsubscript{\textcolor{PineGreen}{$\uparrow$1.30}} & 95.28\textsubscript{\textcolor{PineGreen}{$\uparrow$2.01}} & 94.68\textsubscript{\textcolor{PineGreen}{$\uparrow$2.19}} & 93.26\textsubscript{\textcolor{PineGreen}{$\uparrow$2.22}} & 76.13\textsubscript{\textcolor{PineGreen}{$\uparrow$0.13}} & 74.72\textsubscript{\textcolor{PineGreen}{$\uparrow$0.40}} & 72.32\textsubscript{\textcolor{OrangeRed}{$\downarrow$0.05}} & 70.04\textsubscript{\textcolor{PineGreen}{$\uparrow$0.17}} & 66.73\textsubscript{\textcolor{PineGreen}{$\uparrow$0.47}} \\

TDDS & 95.49\textsubscript{\textcolor{PineGreen}{$\uparrow$0.82}} & 95.49\textsubscript{\textcolor{PineGreen}{$\uparrow$1.34}} & 95.30\textsubscript{\textcolor{PineGreen}{$\uparrow$2.03}} & 94.66\textsubscript{\textcolor{PineGreen}{$\uparrow$2.17}} & 93.51\textsubscript{\textcolor{PineGreen}{$\uparrow$2.47}} & 77.25\textsubscript{\textcolor{PineGreen}{$\uparrow$1.25}} & 76.00\textsubscript{\textcolor{PineGreen}{$\uparrow$1.68}} & 74.09\textsubscript{\textcolor{PineGreen}{$\uparrow$1.72}} & 71.91\textsubscript{\textcolor{PineGreen}{$\uparrow$2.04}} & 68.38\textsubscript{\textcolor{PineGreen}{$\uparrow$2.12}} \\

Moderate & 94.37\textsubscript{\textcolor{OrangeRed}{$\downarrow$0.30}} & 93.89\textsubscript{\textcolor{OrangeRed}{$\downarrow$0.26}} & 93.22\textsubscript{\textcolor{OrangeRed}{$\downarrow$0.05}} & 92.42\textsubscript{\textcolor{OrangeRed}{$\downarrow$0.07}} & 90.86\textsubscript{\textcolor{OrangeRed}{$\downarrow$0.18}} & 76.06\textsubscript{\textcolor{PineGreen}{$\uparrow$0.06}} & 74.74\textsubscript{\textcolor{PineGreen}{$\uparrow$0.42}} & 72.84\textsubscript{\textcolor{PineGreen}{$\uparrow$0.47}} & 70.74\textsubscript{\textcolor{PineGreen}{$\uparrow$0.87}} & 66.38\textsubscript{\textcolor{PineGreen}{$\uparrow$0.12}} \\

\rowcolor{gray!10}
UNSEEN & \textbf{95.59}\textsubscript{\textcolor{PineGreen}{$\uparrow$0.92}} & \textbf{95.58}\textsubscript{\textcolor{PineGreen}{$\uparrow$1.43}} & \textbf{95.35}\textsubscript{\textcolor{PineGreen}{$\uparrow$2.08}} & \textbf{95.10}\textsubscript{\textcolor{PineGreen}{$\uparrow$2.61}} & \textbf{94.16}\textsubscript{\textcolor{PineGreen}{$\uparrow$3.12}} & \textbf{78.61}\textsubscript{\textcolor{PineGreen}{$\uparrow$2.61}} & \textbf{76.88}\textsubscript{\textcolor{PineGreen}{$\uparrow$2.56}} & \textbf{75.15}\textsubscript{\textcolor{PineGreen}{$\uparrow$2.78}} & \textbf{71.99}\textsubscript{\textcolor{PineGreen}{$\uparrow$2.12}} & \textbf{68.49}\textsubscript{\textcolor{PineGreen}{$\uparrow$2.23}} \\
\bottomrule
\end{tabular}
}
\label{CIFAR-table}
\end{table*}

\begin{table}[tb!]
\centering
\caption{Comprehensive Comparison on ImageNet-1K. The accuracy on the full dataset is 73.61\%. UNSEEN prunes 30\% of the data with only a 0.06\% accuracy drop.}
\resizebox{0.6\textwidth}{!}{
\begin{tabular}{@{}c| *{3}{c} @{}}
\toprule
\multicolumn{1}{c|}{\centering Dataset} & \multicolumn{3}{c}{\textbf{ImageNet-1K}} \\
\cmidrule(lr){1-1}\cmidrule(lr){2-4}
Prune Rate & 30\% & 50\% & 70\% \\
\cmidrule(lr){1-1}\cmidrule(lr){2-4}

Random & 72.16 & 71.07 & 70.00  \\
Entropy & 72.72\textsubscript{\textcolor{PineGreen}{$\uparrow$0.56}} & 70.93\textsubscript{\textcolor{OrangeRed}{$\downarrow$0.14}} & 67.55\textsubscript{\textcolor{OrangeRed}{$\downarrow$2.45}} \\
Margin & 72.10\textsubscript{\textcolor{OrangeRed}{$\downarrow$0.06}} & 70.93\textsubscript{\textcolor{OrangeRed}{$\downarrow$0.14}} & 67.58\textsubscript{\textcolor{OrangeRed}{$\downarrow$2.42}} \\
Least Confidence & 72.32\textsubscript{\textcolor{PineGreen}{$\uparrow$0.16}} & 71.18\textsubscript{\textcolor{PineGreen}{$\uparrow$0.11}} & 67.47\textsubscript{\textcolor{OrangeRed}{$\downarrow$2.53}} \\
AUM & 72.96\textsubscript{\textcolor{PineGreen}{$\uparrow$0.80}} & 67.47\textsubscript{\textcolor{OrangeRed}{$\downarrow$3.60}} & 44.58\textsubscript{\textcolor{OrangeRed}{$\downarrow$25.42}} \\
EL2N & 72.65\textsubscript{\textcolor{PineGreen}{$\uparrow$0.49}} & 69.61\textsubscript{\textcolor{OrangeRed}{$\downarrow$1.46}} & 62.78\textsubscript{\textcolor{OrangeRed}{$\downarrow$7.22}} \\
Forgetting & 71.96\textsubscript{\textcolor{OrangeRed}{$\downarrow$0.20}} & 70.26\textsubscript{\textcolor{OrangeRed}{$\downarrow$0.81}} & 68.14\textsubscript{\textcolor{OrangeRed}{$\downarrow$1.86}} \\
Moderate & 72.47\textsubscript{\textcolor{PineGreen}{$\uparrow$0.31}} & 70.94\textsubscript{\textcolor{OrangeRed}{$\downarrow$0.13}} & 67.42\textsubscript{\textcolor{OrangeRed}{$\downarrow$2.58}} \\
CCS & 71.65\textsubscript{\textcolor{OrangeRed}{$\downarrow$0.51}} & 70.09\textsubscript{\textcolor{OrangeRed}{$\downarrow$0.98}} & 66.30\textsubscript{\textcolor{OrangeRed}{$\downarrow$3.70}} \\
DynUnc & 70.28\textsubscript{\textcolor{OrangeRed}{$\downarrow$1.88}} & 66.39\textsubscript{\textcolor{OrangeRed}{$\downarrow$4.68}} & 60.69\textsubscript{\textcolor{OrangeRed}{$\downarrow$9.31}} \\
TDDS & 71.47\textsubscript{\textcolor{OrangeRed}{$\downarrow$0.69}} & 68.91\textsubscript{\textcolor{OrangeRed}{$\downarrow$2.16}} & 64.52\textsubscript{\textcolor{OrangeRed}{$\downarrow$5.48}} \\
\rowcolor{gray!10}
UNSEEN & \textbf{73.55}\textsubscript{\textcolor{PineGreen}{$\uparrow$1.39}} & \textbf{72.13}\textsubscript{\textcolor{PineGreen}{$\uparrow$1.06}} & \textbf{70.26}\textsubscript{\textcolor{PineGreen}{$\uparrow$0.26}} \\
\bottomrule
\end{tabular}
}
\label{ImageNet-table}
\end{table}

\section{Experiment}

\subsection{Dataset and Settings}
 \noindent \textbf{Datasets and architecture.} We evaluated our framework UNSEEN on CIFAR-10/100~\citep{krizhevsky2009learning} and ImageNet-1K~\citep{deng2009imagenet}. We further evaluated our approach on three challenging fine-grained visual categorization (FGVC) datasets: CUB-2011~\citep{wah2011caltech}, Stanford Dogs~\citep{khosla2011novel}, and Stanford Cars~\citep{krause20133d}. To validate cross-architecture generalization, we implement dataset pruning on CIFAR-10 and CIFAR-100 using ResNet-18~\citep{he2016deep} as the backbone, followed by evaluation of coreset transfer performance on ResNet-34 and ResNet-50.

 \noindent \textbf{Baseline.} We compare our approach against eleven baselines, including Random, Entropy~\citep{coleman2019selection}, Margin~\citep{coleman2019selection}, Least Confidence~\citep{coleman2019selection},  EL2N~\citep{paul2021deep}, AUM~\citep{pleiss2020identifying}, Forgetting~\citep{toneva2018an}, Moderate~\citep{xia2022moderate}, CCS~\citep{zheng2022coverage}, DynUnc~\citep{he2024large}, TDDS~\citep{zhang2024spanning}. 

\textbf{Implementation details.}
We set $K = 4$ and $J = 2$, \emph{i.e.}, only one additional pruning stage, with $M_2 = \left\lceil N \cdot 10\% \right\rceil$ as the default, consistent across all pruning rates. All experimental results are averaged over five runs. Other details are in the supplementary materials.

\subsection{Experiments Results}

\textbf{\noindent{Performance of general image classification}}. UNSEEN outperformed previous dataset pruning methods on CIFAR-10, CIFAR-100, and ImageNet-1K at varying pruning rates. As shown in \Cref{CIFAR-table,ImageNet-table}, UNSEEN achieved 73.55\% accuracy when pruning 30\% samples of ImageNet-1K, with only a 0.06\% accuracy gap compared to training on full data.

\begin{figure}[tb!]
  \centering
  \includegraphics[width=0.7\linewidth]{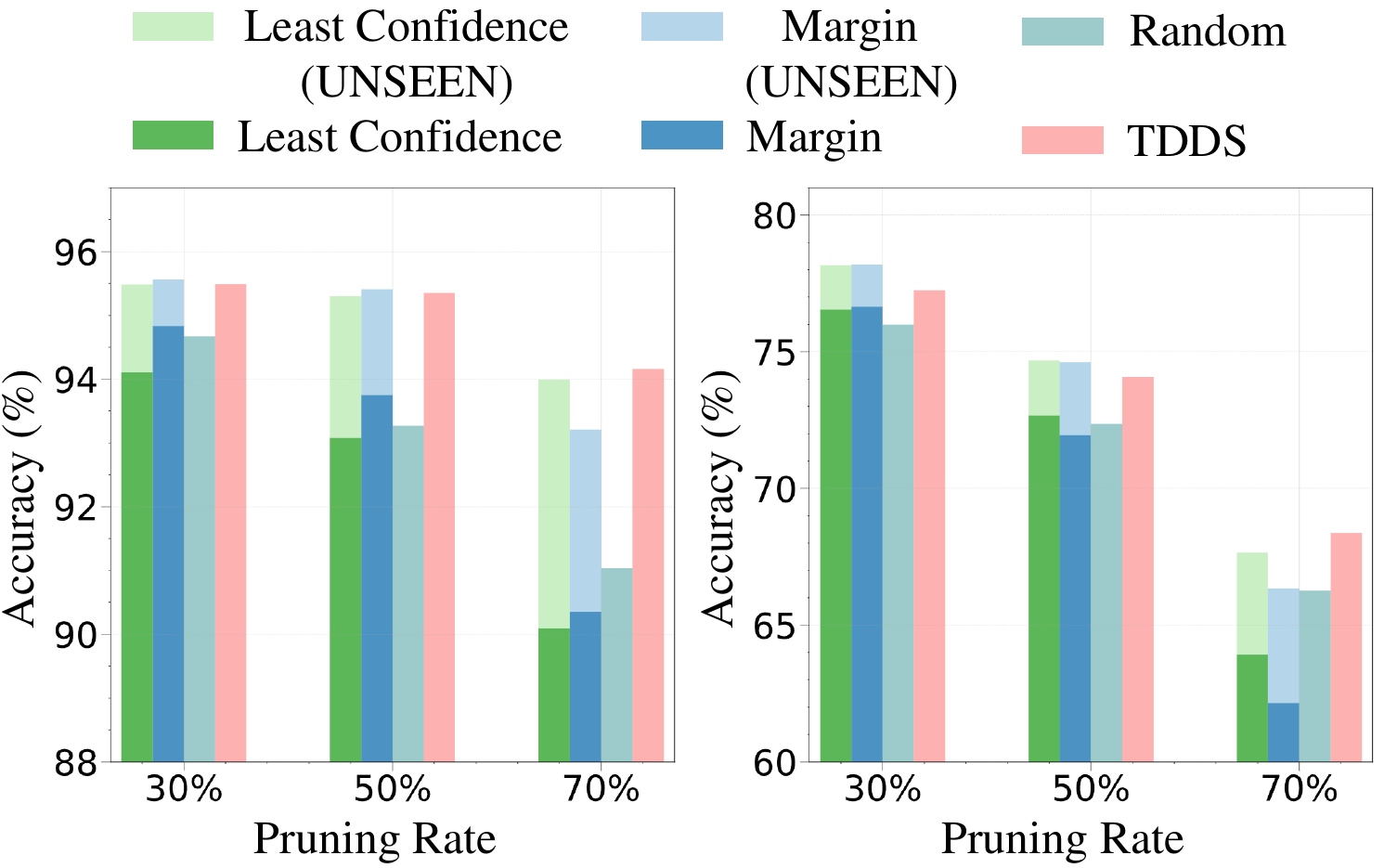}
  \caption{Plug-and-play enhancement of UNSEEN on CIFAR-10 (left) and CIFAR-100 (right). Margin and Least Confidence achieve significant enhancement with UNSEEN and outperform TDDS at low pruning rates.}
  \label{plug-and-play-fig}
\end{figure}

\noindent \textbf{UNSEEN can be applied to existing methods}: 
\Cref{plug-and-play-fig} showed the performance of random dataset pruning, two classical methods (Margin and Least Confidence), and the previous SOTA method TDDS. We applied UNSEEN to the two classical methods. With UNSEEN, they significantly outperformed baselines and approached TDDS performance. For example, Margin with UNSEEN and Least Confidence with UNSEEN achieve accuracy of 78.19\%  and 78.17\% at a pruning ratio of 30\% on CIFAR-100, respectively, demonstrating that UNSEEN is a plug-and-play framework that can be incorporated into existing methods.

\begin{figure}[tb!]
  \centering
  \includegraphics[width=0.7\linewidth]{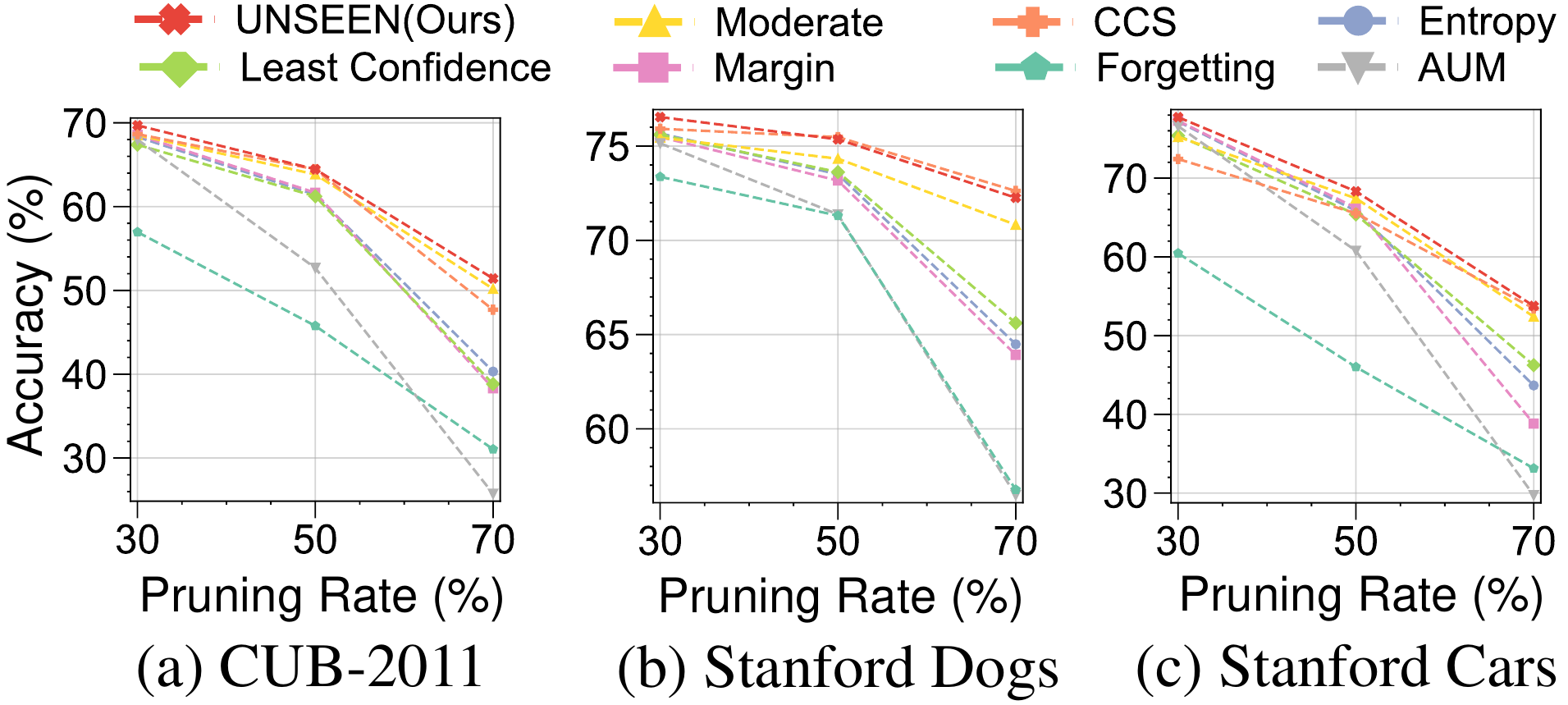}

  \caption{Comprehensive comparison on fine-grained datasets, demonstrating UNSEEN’s superior performance.}
  \label{fine-grained}
\end{figure}

\noindent \textbf{\textbf{Performance of fine-grained image classification}}. We also applied UNSEEN to datasets with subtle image differences. We conduct experiments on CUB-2011, Stanford Dogs, and Stanford Cars. As illustrated in \Cref{fine-grained}, our method constructs informative coresets during fine-tuning across all three datasets, demonstrating UNSEEN’s effectiveness in selecting fine-grained samples with minimal variations.

\begin{figure*}[tb!]
  \centering
  \includegraphics[width=\linewidth]{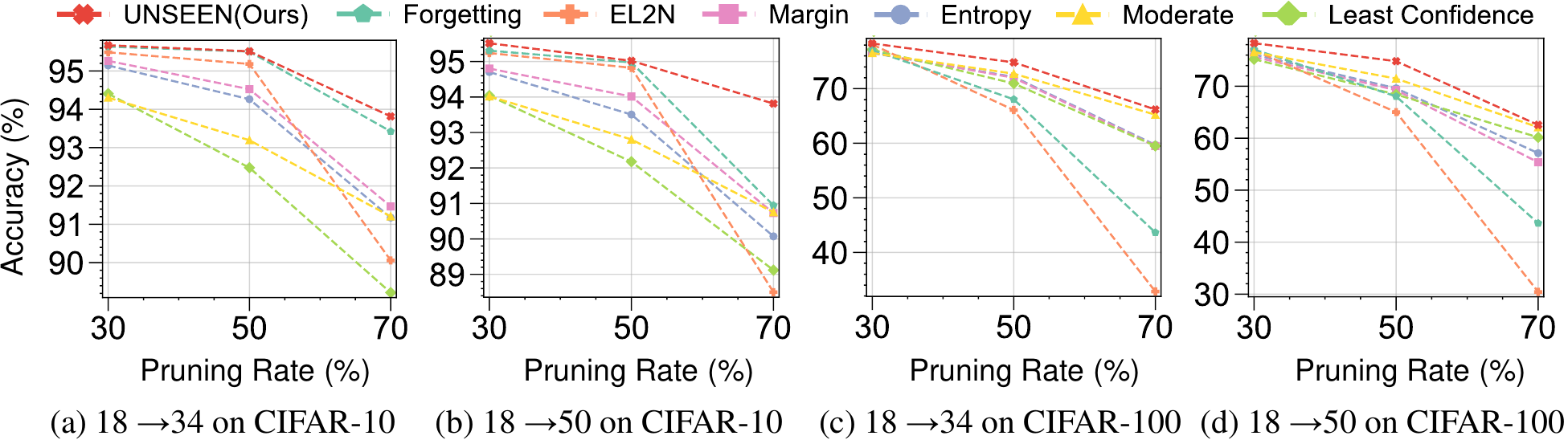} 
  \caption{We employed ResNet-18 to perform dataset pruning, and subsequently trained ResNet-34 and ResNet-50 on the pruned datasets. Results demonstrate that the coreset selected by UNSEEN exhibits strong generalization across architectures.}
  \label{cross-architecture generalization}
\end{figure*}

\noindent \textbf{Cross-architecture generalization}. 
We validated the applicability of pruned datasets on larger, unseen network architectures not involved in pruning. Specifically, we pruned CIFAR-10 and CIFAR-100 with ResNet-18, and trained the obtained coreset with ResNet-34 and ResNet-50. As shown in \Cref{cross-architecture generalization}, our method significantly surpasses other methods, demonstrating that our method generates a coreset with remarkable cross-architecture generalization capabilities.

\subsection{Ablation study}
\noindent \textbf{Ablation study on UNSEEN and IS}.

We adopted Entropy as the baseline, then integrated UNSEEN and IS separately. As shown in~\Cref{ablation unseen-is}, they both exhibit significant improvements, with UNSEEN demonstrating a more pronounced enhancement. The optimal result is achieved when combined.

\begin{table}[tb!]
\setlength{\tabcolsep}{4pt} %
\centering
\footnotesize 
\begin{tabular}{ccccc|ccc}
\toprule
\multicolumn{2}{c}{\textbf{Dataset}} & \multicolumn{3}{c|}{\textbf{CIFAR-10}} & \multicolumn{3}{c}{\textbf{CIFAR-100}} \\
\cmidrule(lr){1-2} \cmidrule(lr){3-5} \cmidrule(lr){6-8}
UNSEEN & IS & 30\% & 50\% & 70\% & 30\% & 50\% & 70\% \\
\midrule
\texttimes & \texttimes & 94.77 & 93.87 & 90.83 & 76.73 & 72.21 & 62.47 \\
\checkmark & \texttimes & 95.39 & 95.33 & 94.05 & 78.24 & 74.56 & 62.23 \\
\texttimes & \checkmark & 95.27 & 94.82 & 92.96 & 76.92 & 74.31 & 65.73 \\
\rowcolor{gray!10}
\checkmark & \checkmark & \textbf{95.59} & \textbf{95.35} & \textbf{94.16} & \textbf{78.61} & \textbf{75.15} & \textbf{68.49} \\
\bottomrule
\end{tabular}

\caption{Ablation study on UNSEEN and Incremental Selection (IS). Both UNSEEN and IS improve the Entropy method, with the optimal results when combined.}
\label{ablation unseen-is}
\end{table}

\noindent \textbf{Ablation study on the number of partitions $K$}. Results showed that models with moderate capability achieve optimal pruning. As analyzed in \Cref{ablation K}, overly strong models (\emph{e.g.},  with extremely small \textit{K}) tend to fit samples precisely, reducing the discrimination for the difficulty of samples. Weaker models (\emph{e.g.}, with excessively large \textit{K}) struggle to effectively capture fundamental class characteristics, resulting in suboptimal differentiation.

\begin{table}[tb!]
\setlength{\tabcolsep}{3pt} %
\centering
\footnotesize 
\begin{tabular}{@{}c*{6}{c}@{}}
\toprule
\multicolumn{1}{c}{Dataset}
 & \multicolumn{3}{c}{\textbf{CIFAR-10}} & \multicolumn{3}{c}{\textbf{CIFAR-100}} \\
\cmidrule(lr){2-4} \cmidrule(lr){5-7}
Prune Rate & 30\% & 50\% & 70\% & 30\% & 50\% & 70\% \\
\midrule
UNSEEN ($K = 2$) & 95.37 & 94.81 & 92.82 & 77.87 & 74.51 & 64.53 \\
\rowcolor{gray!10}
UNSEEN ($K = 4$) & \textbf{95.39} & \textbf{95.34} & \textbf{94.05} & \textbf{78.24} & \textbf{74.56} & \textbf{67.90} \\
UNSEEN ($K = 10$) & 95.37 & 95.24 & 93.28 & 76.98 & 73.45 & 67.52 \\
UNSEEN ($K = 20$) & 95.29 & 94.51 & 92.89 & 76.23 & 72.11 & 66.41 \\
\bottomrule
\end{tabular}

\caption{Ablation study on the number of folders $K$. UNSEEN yields optimal performance for moderate $K$ values.}
\label{ablation K}
\end{table}

\noindent \textbf{Abation study on generalization-based scoring}. Scoring models of UNSEEN can be viewed as proxy models trained on reduced data. We conducted experiments to validate the effectiveness of UNSEEN from a generalization perspective. For the proxy model based on Entropy, the dataset was randomly partitioned into $K$ equally sized subsets, and $K$ scoring models were trained on these subsets to assign scores to samples within each subset (\emph{i.e.,} Entropy-proxy). We set \textit{K} = 4, the same as the experimental setting in UNSEEN.  \Cref{ablation subset-based} compares the results of dataset pruning among the original Entropy, Entropy-proxy, and UNSEEN. UNSEEN significantly outperforms Entropy-proxy, demonstrating its advantages from a generalization perspective.

\begin{table}[tb!]
\setlength{\tabcolsep}{2pt}
\centering
\footnotesize 
\begin{tabular}{@{}c*{6}{c}@{}}
\toprule
\multicolumn{1}{c}{Dataset}
 & \multicolumn{3}{c}{\textbf{CIFAR-10}} 
 & \multicolumn{3}{c}{\textbf{CIFAR-100}} \\
\cmidrule(lr){2-4} \cmidrule(lr){5-7}
Prune Rate & 30\% & 50\% & 70\% & 30\% & 50\% & 70\% \\
\midrule
Entropy & 94.77 & 93.87 & 90.83 & 76.73 & 72.21 & 62.47 \\
Entropy-proxy ($K=4$) & 94.85 & 94.04 & 92.04 & 76.83 & 73.22 & 65.08 \\
\rowcolor{gray!10}
UNSEEN ($K=4$) & \textbf{95.59} & \textbf{95.35} & \textbf{94.16} & \textbf{78.61} & \textbf{75.15} & \textbf{67.90} \\
\bottomrule
\end{tabular}

\caption{Ablation study on generalization-based scoring. UNSEEN significantly outperforms Entropy-proxy, demonstrating its advantages from a generalization perspective.}
\label{ablation subset-based}
\end{table}

\begin{figure}[tb!]
    \centering
    \begin{subfigure}[b]{0.48\textwidth}
        \centering
        \includegraphics[width=\textwidth]{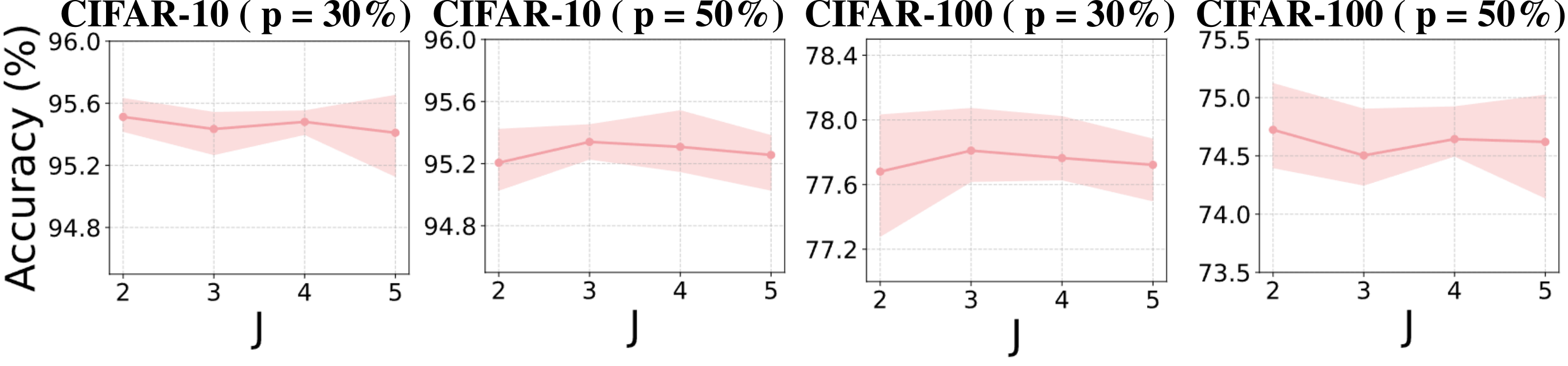}
        \caption{Ablation Study of $J$}
        \label{fig:ablation_J}
    \end{subfigure}
    \hfill
    \begin{subfigure}[b]{0.48\textwidth}
        \centering
        \includegraphics[width=\textwidth]{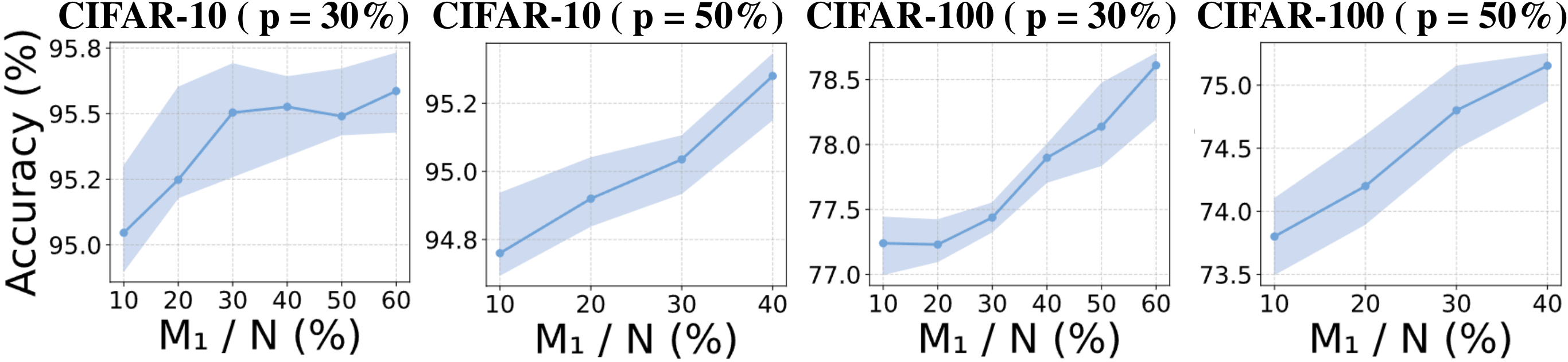}
        \caption{Ablation Study of $M$ ($J$ = 2)}
        \label{fig:ablation_M}
    \end{subfigure}
    \caption{Ablation studies of hyperparameters $J$ and $M$.}
    \label{fig:ablation_joint}
\end{figure}

\noindent \textbf{Ablation Study for the Number of Refinement Stages $J$ and Per-Stage Coreset Size $M$}.
As shown in \Cref{fig:ablation_J}, although the cost grows with increasing $J$, the performance remains nearly unchanged. We also conducted the ablation study on the first-stage coreset size $M_1$ ($\left\lceil N \cdot (1 - p) \right\rceil - M_2$) using CIFAR-10 and CIFAR-100 when $J = 2$. \Cref{fig:ablation_M} shows that a larger first-stage coreset, supplemented by a smaller second-stage retraining, yields better performance.

\section{Discussion}
\subsection{Computational Cost of UNSEEN and IS}
\textbf{UNSEEN does not introduce any additional cost, while incremental selection incurs only minimal overhead.} \\
(i) Computational Overhead of Previous Methods: Given a batch size of $B$, previous methods incur $N/B$ iterations to train a scoring model on the full dataset. After pruning $p$ of the data, the model is trained on the selected coreset with $N \cdot (1 - p)/B$ iterations. Thus, the total computational overhead of previous methods is $N \cdot (2 - p)/B$ iterations. \\ (ii) Computational Overhead of UNSEEN: UNSEEN trains $K$ scoring models on $K$ equal-sized subsets. Each requires $N/(K \cdot B)$ iterations, summing to $N/B$ iterations. Training on the coreset adds $N \cdot (1 - p)/B$ iterations, totaling $N \cdot (2 - p)/B$ iterations, which is \textit{equivalent to that of previous methods}. Hence, \textbf{UNSEEN does not introduce any additional computational cost.} \\
(iii) Computational Overhead of Incremental Selection (IS): We set $J=2$ and 
$M_2 = \left\lceil N \cdot 10\% \right\rceil$
, \emph{i.e.}, $10\%$ of the full dataset is incrementally selected in the second step. An additional model is trained before the second stage, incurring $N (1 - p - 10\%)/B$ iterations. For instance, \textbf{when pruning $70\%$ of the data, IS incurs only $0.15\times$ the total cost required by previous methods}~(see figure below).

\begin{figure}[htbp]
    \centering
    \includegraphics[width=0.7\textwidth]{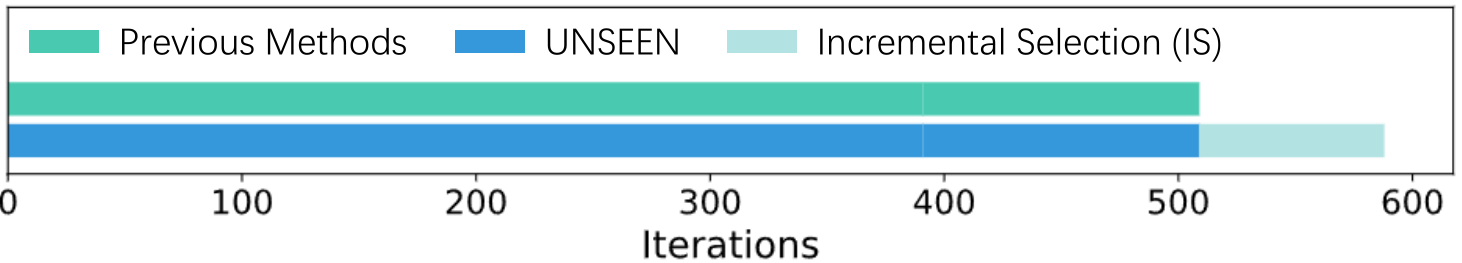}
    \label{Iteration}
    \caption{Comparison of Computational Overhead among Previous Methods, UNSEEN, and IS at 70\% Pruning Rate.}
\end{figure}

\begin{figure}[tb!]
  \centering
  \begin{subfigure}{0.3\linewidth}
    \includegraphics[width=\textwidth]{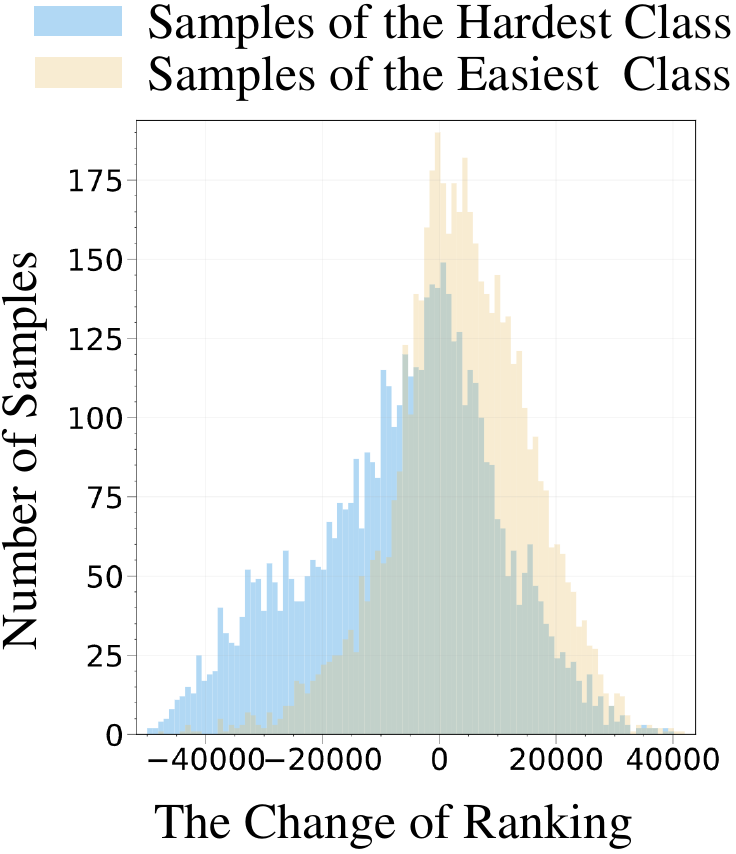}
    \caption{Ranking Change}
    \label{fig:rank_change} 
  \end{subfigure}
  \begin{subfigure}{0.3\linewidth}
    \includegraphics[width=\textwidth]{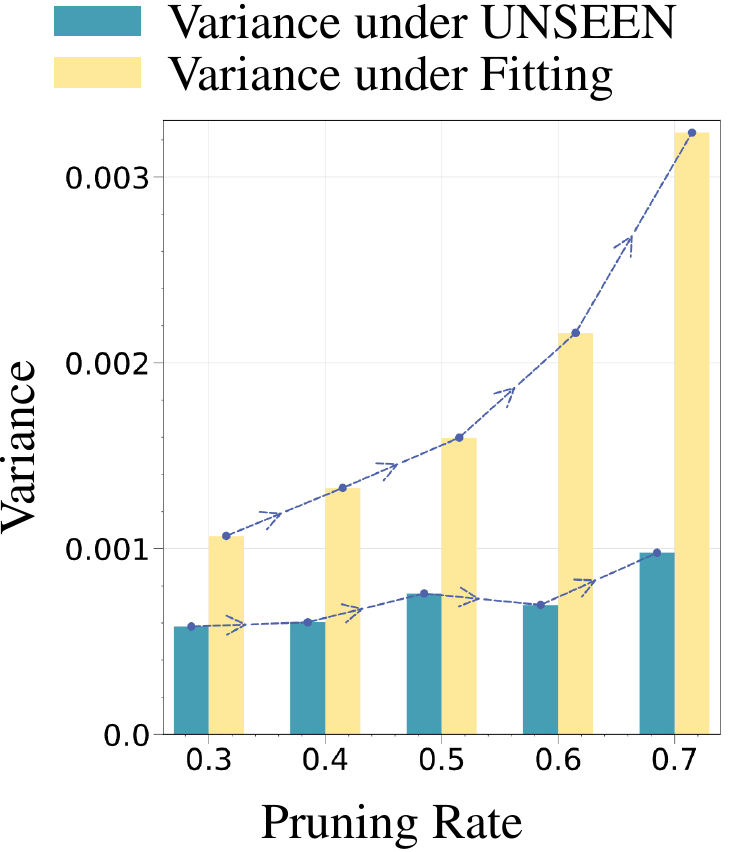}
    \caption{Inter-class Var Comparison}
    \label{fig:class_var}
  \end{subfigure}
  
  \caption{ 
    (a) We computed the differential ranking by subtracting the fitting rankings from the UNSEEN rankings. We observed that most samples in the hardest class experienced rank drops (prior to being selected in UNSEEN), while the easiest class exhibited the opposite trend. 
    (b) We measured variance in classification accuracies across classes (\emph{i.e.,} inter-class variance). With increasing pruning rates, the Entropy method under the fitting framework showed rapid growth in inter-class variance, whereas UNSEEN maintained significantly stable and lower inter-class variance.
  }

  \label{fig:discussion}
\end{figure}

\subsection{Class-Level Analysis of UNSEEN}
This section explains UNSEEN’s strong performance by analyzing dataset pruning at the class level. Given that different classes inherently possess varying levels of complexity, models tend to prioritize learning simpler classes, leading to a lower accuracy for challenging ones and consequently creating an imbalance in performance~\citep{cui2024classes}. Most existing pruning methods either ignore class information or treat all classes uniformly~\citep{guo2022deepcore} (\emph{i.e.,} selecting the same number of samples from each class).

Since the scoring models under UNSEEN are not exposed to training samples, they implicitly incorporate amplified class-level difficulty weighting, which leads to a prioritization of hard-class samples. To verify this hypothesis, we selected all samples from the easiest and the hardest classes in the full dataset and plotted the change in their score rankings when transitioning from the fitting framework to UNSEEN (i.e., the difference in rankings between the two frameworks). Lower rankings indicate higher scores and greater selection priority. As demonstrated in ~\Cref{fig:rank_change}, a majority of samples from the hardest class exhibited negative differences with lower magnitudes, indicating their increased selection priority under the UNSEEN framework. In contrast, samples from the easiest class exhibited a contrasting pattern with positive differences. The prioritization of hard-class instances facilitates mitigating inter-class discrepancy within the coreset, thereby achieving holistic performance optimization through balanced representation learning. As shown in \Cref{fig:class_var}, we observe that with increasing pruning rates, the inter-class variance \emph{i.e.,} variance of accuracies across different classes, increases rapidly under the fitting framework, while the inter-class variance is stably lower under the UNSEEN framework. The experimental result demonstrates that minimizing the accuracy disparity between classes can enhance overall performance.

\section{Conclusion}
In this paper, we identify that existing dataset pruning methods under the fitting framework yield highly dense scores, leading to undiscriminating and unstable selection.  Therefore, we introduce a plug-and-play framework, UNSEEN, from the perspective of generalization. To refine the previous single-step pruning method, we scale UNSEEN to multi-step scenarios and propose incremental selection to evaluate samples comprehensively and optimize the coreset dynamically. Finally, we analyze the reason for UNSEEN's outstanding performance by prioritizing samples of hard classes and extend the concept of difficulty from samples to classes. Experiments demonstrate that minimizing inter-class disparity is critical for achieving exceptional performance.

\bibliography{colm2024_conference}
\bibliographystyle{colm2024_conference}

\end{document}